\pdfoutput=1

\documentclass[a4paper,11pt]{article}

\usepackage[preprint]{acl}

\usepackage[T1]{fontenc}
\usepackage[utf8]{inputenc}

\usepackage{times}

\usepackage{amsmath}
\usepackage{amsthm}
\usepackage{amssymb}

\usepackage{latexsym}

\usepackage{microtype} 
\usepackage{inconsolata} 

\usepackage{subfiles}

\usepackage[most]{tcolorbox}
\usepackage{booktabs}
\usepackage{multirow}
\usepackage{enumitem}
\usepackage{graphicx}
\usepackage{xcolor}
\usepackage{hyperref}
\usepackage[capitalise]{cleveref}
\usepackage{caption}

\usepackage{siunitx} 
\usepackage{CJKutf8} 
\usepackage[safe]{tipa} 
\usepackage{comment}
\usepackage{csquotes} 
\usepackage{twemojis}

\newcommand{\zh}[1]{\begin{CJK}{UTF8}{gbsn}#1\end{CJK}}

\definecolor{myblue}{RGB}{20,80,150}  
\definecolor{myred}{RGB}{160,30,30}   
\definecolor{mygreen}{RGB}{50,120,50} 

\title{What Makes a Word Hard to Learn?\\Modeling L1 Influence on English Vocabulary Difficulty}

\author{
    \bf{Jonas Mayer Martins} ~~~~
    Zhuojing Huang ~~~~
    Aaricia Herygers ~~~~
    Lisa Beinborn \\
    University of G\"{o}ttingen, Germany \\ 
    \texttt{firstname.lastname@uni-goettingen.de} \\}
\begin{document}

\maketitle

\begin{abstract}
What makes a word difficult to learn, and how does the difficulty depend on the learner's native language? We computationally model vocabulary difficulty for English learners whose first language is Spanish, German, or Chinese with gradient-boosted models trained on features related to a word's familiarity (e.g., frequency), meaning, surface form, and cross-linguistic transfer. Using Shapley values, we determine the importance of each feature group. Word familiarity is the dominant feature group shared by all three languages. However, predictions for Spanish- and German-speaking learners rely additionally on orthographic transfer. This transfer mechanism is unavailable to Chinese learners, whose difficulty is shaped by a combination of familiarity and surface features alone. Our models provide interpretable, L1-tailored difficulty estimates that can be used to design vocabulary curricula.
\end{abstract}

\begin{center}
	\small
	{
		\raisebox{-0.33em}{\includegraphics[height=1.2em]{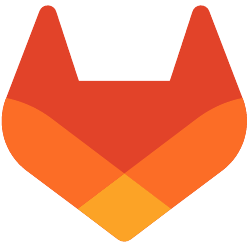}} 
		\href{https://gitlab.gwdg.de/huds/projects/bea2026/-/tree/v1.0.2}{Code repository} and 
		\href{https://huds.pages.gwdg.de/projects/bea2026}{interactive demo}
	}
\end{center}


\section{Introduction} 

Learning a second language begins with words. Developing an extensive vocabulary is essential for mastering grammar and achieving fluency \citep{schmitt-schmitt-2020-vocabulary,nation-2000-learning}. Yet not all words are equally hard to learn, and researchers of second-language (L2) acquisition, classroom practitioners, and creators of educational materials seek to better understand this phenomenon.

The difficulty of a word is partially due to factors common to all learners, that is, lexical properties that depend only on the target language itself \citep{peters-2019-factors}. For example, common and concrete words are easier to acquire than rare and abstract ones \citep{ellis-2002-frequency,ellis-beaton-1993-psycholinguistic}.

However, a learner's individual background---their first language (L1), in particular---plays a key role in word difficulty \citep{ringbom-jarvis-2009-importance}. For instance, a German speaker readily infers that the English word \enquote{sheep} is related to the cognate \emph{Schaf}, whereas the Spanish \emph{oveja} is not a cognate.\footnote{Instead, \emph{oveja} is distantly related to \enquote{ewe}, meaning female sheep, in English.} This orthographic bridge supports active as well as passive vocabulary competence. Conversely, false friends may lead to interfering transfer \citep{bensoussan-laufer-1984-lexical}. For example, \emph{bravo} in Spanish is close to its English cognate \enquote{brave}, yet \emph{brav} in German means \enquote{well-behaved}. The same English word can thus be easy for one learner while difficult for another, depending on the relationship between the L1 and~L2.

However, computational models of lexical difficulty often fail to take a learner's background into account. The SemEval2021 shared task on lexical-complexity prediction, for example, considers difficulty for unspecified annotators \citep{shardlow-etal-2021-semeval2021}. Meanwhile, predictions of lexical difficulty tailored to individual learners are receiving increasing attention \citep{palmeroaprosio-etal-2020-adaptive,north-etal-2023-lexical,schmitt-etal-2021-introducing,skidmore-etal-2025-transformer}, yet studies that do include L1 effects often remain limited to a single language pair.

\paragraph{Approach and contributions.}

\begin{figure*}[tbhp]
    \centering
    \includegraphics[width=\linewidth]{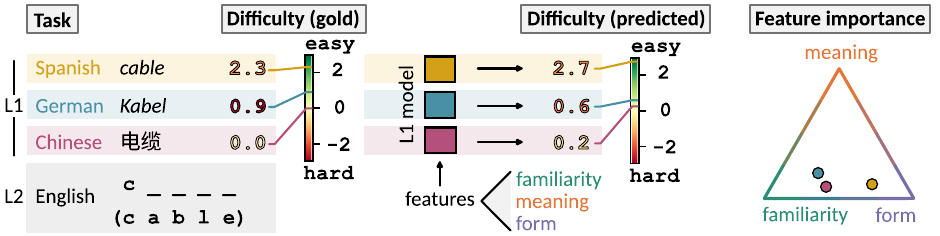}
    \caption{Illustration of the task and modeling setup. For each L1, learners translate a word (e.g., \texttt{Kabel}) from their L1 into English given the first letter (\texttt{c}) and a graphical hint indicating the number of characters. Item difficulty is estimated from aggregated responses (gold label). We train L1-specific models on feature groups (familiarity, meaning, and form = surface $\cup$ transfer). By evaluating the feature importance, we assign a relative importance in the prediction of a word to each feature group.}
    \label{fig:graphical_abstract}
\end{figure*}

We address this gap by modeling English vocabulary difficulty separately for Spanish, German, and Chinese\footnote{In this article, we refer to Mandarin Chinese as Chinese.} L1 speakers, using the Knowledge-based Vocabulary Lists (KVL), a recent crowd-sourced corpus of lexical-difficulty tests \citep{schmitt-etal-2021-introducing,skidmore-etal-2025-transformer}, in the context of the closed track of the Building Educational Applications (BEA) 2026 shared task \citep{felice-skidmore-2026-shared}. We create four feature groups: the familiarity of a word based on exposure effects, its meaning based on semantic complexity and concreteness, its surface form based on orthographic complexity, and its transferability from a specific L1 based on string similarity. We train interpretable, gradient-boosted models on these features to predict L1-dependent lexical difficulty and trace their predictive power via Shapley values. \Cref{fig:graphical_abstract} illustrates the task format and our approach.

To showcase our results and support further research and practical use, we have created an \href{https://huds.pages.gwdg.de/projects/bea2026}{interactive demo} that extends our predictions to more than 4{,}000 to 7{,}000 additional English words per language, see \cref{sec:app:demo} and \cref{sec:app:KVL_extension}. This demo highlights the advantage of our computational modeling approach, which allows us to make predictions beyond the limited KVL data for a larger range of vocabulary and potentially more L1 backgrounds, providing tailored difficulty estimates applicable to curriculum design.


\section{Related work}
\label{sec:related-work}

In this section, we discuss factors of word difficulty in a foreign language based on second-language-learning studies and computational models.

\subsection{Lexical difficulty}

Vocabulary knowledge poses a fundamental challenge to all second-language (L2) learners \citep{schmitt-schmitt-2020-vocabulary,nation-2000-learning}. However, the difficulty associated with a particular word varies by learner. Some words, for instance, represent intrinsically difficult concepts in the target language, while others are challenging or easy specifically due to a learner's L1 background.

\paragraph{L2-intrinsic difficulty.} 
The more frequent a stimulus occurs, the more readily it tends to be recalled, and as such, frequent words are learned more easily because learners encounter them more often \citep{ellis-2002-frequency,peters-2019-factors}. Educational curricula typically introduce frequent words first to enable learners to speak and understand the target language earlier by efficiently building vocabulary coverage \citep{nation-2000-learning}.

Beyond frequency, other lexical properties are also important. For instance, orthographic complexity has been shown to predict exercise difficulty \citep{beinborn-etal-2016-predicting}. Additionally, homonymous or polysemous words are typically harder to process and acquire than words with just one meaning \citep{bensoussan-laufer-1984-lexical}.
Psycholinguistic features, such as the age of acquisition \citep{kuperman-etal-2012-ageofacquisition,dascalu-etal-2016-age} and concreteness of the word \citep{ellis-beaton-1993-psycholinguistic, vanhell-mahn-1997-keyword, degroot-keijzer-2000-what}, are important factors for word difficulty prediction, too. The initial letter plays a disproportionate role in word recognition and production: speakers in tip-of-the-tongue states often remember the first letter and phonological primes aid retrieval \citep{brown-mcneill-1966-tip,james-burke-2000-phonological}.

\paragraph{L1-specific difficulty.} 
Beyond L2-intrinsic word properties, difficulty is shaped by the learner's first language. Learners draw on L1 knowledge when acquiring an L2 vocabulary, with positive transfer where the languages are similar and interference where surface similarity is misleading \citep{odlin-1989-language}. At the lexical level, cognateness is a well-studied transfer mechanism. Cognates reduce learning effort because learners can map the L2 target onto the well-known L1 form: \citet{degroot-keijzer-2000-what} show that cognateness predicts L2-vocabulary retention, and \citet{beinborn-etal-2014-readability} find that learners infer the meaning of unfamiliar words more readily when there is a cognate in their~L1. Similarly, \citet{urdaniz-skoufaki-2022-spanish} find that frequency and cognateness are the two strongest predictors of lexical difficulty for Spanish learners of English, with significant interaction such that word frequency is more predictive for non-cognates than cognates. However, coincidental similarity or shifted meaning of cognates can cause interference \citep{odlin-1989-language,bensoussan-laufer-1984-lexical}. 

These findings imply that the same word can be easy for one learner and hard for another, depending on the overlap of L1 and L2. Moreover, cognateness interacts with other predictors of difficulty, motivating models that jointly incorporate L1-specific and L2-intrinsic information. 

\subsection{Predicting lexical difficulty}

Progress on predicting lexical difficulty\footnote{In this article, we use \textit{difficulty} to refer to the processing effort in retrieval, as defined in \citet{bulte-etal-2025-complexity}, and distinguish this from structural complexity, which concerns the internal composition of a word.} has been driven by shared-task competitions \citep{paetzold-specia-2016-semeval,yimam-etal-2018-report,shardlow-etal-2021-semeval2021}. Computational approaches to predicting lexical difficulty have shifted from binary classification of complex words to continuous regression of difficulty scores \citep{north-zampieri-2023-features}. Across studies, robust predictors include word frequency, word length, age of acquisition, and concreteness \citep{north-etal-2023-lexical}. 

These patterns extend beyond English.
The CWI-2018 shared task shows that models trained on English, German, and Spanish can generalize to French, suggesting that certain cognitive mechanisms transfer across typologically related languages \citep{yimam-etal-2018-report}. The BEA 2024 shared task investigated lexical-complexity prediction and simplification with 10 languages, finding that feature-based systems can compete with large language models \citep{shardlowetal-2024-bea}. For some languages, script-specific features are important, e.g., logographic-character frequency in Chinese and Japanese \citep{lee-yeung-2018-automatic,nishihara-kajiwara-2020-word}. Furthermore, \citet{finnimore-etal-2019-strong} confirm features from typologically related languages can be beneficial, while including features from unrelated languages may reduce performance. 

Despite these advances, learner-specific factors remain underexplored. Studies that model the L1 background report benefits over agnostic models, e.g., for Dutch learners of French as well as Chinese and Spanish learners of English \citep{tack-etal-2016-modeles,lee-yeung-2018-personalizing,urdaniz-skoufaki-2022-spanish}. However, these studies focus on a single language pair. \citet{palmeroaprosio-etal-2020-adaptive} extend this line of research to multiple L1s for learners of Italian by modeling cognates and false friends. Similarly, the lexical complexity for learners of Japanese has been studied for several L1 backgrounds \citep{ide-etal-2023-japanese,nohejl-etal-2024-difficult}. Still, the interaction between L1-specific and general features remains largely unexamined. In this article, we address that gap directly.


\section{Experimental setup}

We train and evaluate L1-specific CatBoost models \citep{prokhorenkova-etal-2018-catboost} for predicting the difficulty of English vocabulary items. This section describes the dataset, feature groups, model, and evaluation metrics.

\subsection{Data}
\label{sec:data}

We use the Knowledge-based Vocabulary Lists (KVL), a large-scale dataset of lexical difficulty based on vocabulary test responses from more than 100{,}000 second-language learners of English \citep{schmitt-etal-2021-introducing,skidmore-etal-2025-transformer} as part of the BEA 2026 shared task \citep{felice-skidmore-2026-shared}. The data covers three L1s---Spanish, German, and Chinese---with 6{,}091 training, 677 development, and 748 test items per L1.

In the KVL test format, learners see a word in their L1 with a context sentence and must type the English translation, given the first letter and blanks for the remaining letters:
\begin{tcolorbox}[
  colback=teal!5, colframe=teal!60, boxrule=0.8pt,
  left=4pt, right=4pt, top=3pt, bottom=3pt
]
\emph{Kabel} ~~{\small\textcolor{gray}{German, noun}}\\
\emph{\small Achtung, stolpere nicht über das Kabel am Boden.}\\
\texttt{c\,\_\,\_\,\_\,\_}
\end{tcolorbox}
\noindent Each item is scored as correct or incorrect. In the example above, the correct completion would be \texttt{a\,b\,l\,e}. Responses are aggregated into a continuous logarithmic difficulty score by a Rasch model \citep{deboeck-2008-random,dunn-2024-randomitem}, which jointly estimates item difficulty and learner ability such that the resulting centered score reflects the intrinsic difficulty of a word (higher scores are easier).\footnote{Note that the values across languages are not directly comparable in an absolute way but rather as relative to the average word difficulty of an L1.}

The difficulty estimates may vary by L1, see \cref{fig:graphical_abstract}. The word \texttt{cable} is easiest for Spanish speakers (score 2.3) because the corresponding Spanish form is spelled identically; moderately easy for German speakers (score 0.9), who benefit from the German cognate \emph{Kabel}; and hardest for Chinese speakers (score 0.0, \zh{电缆} \textit{di\`{a}nl\v{a}n}).

\subsection{Features}
\label{sec:features}

We extract 24 features per item and organize them into four feature groups that structure our analysis, see \cref{fig:graphical_abstract}. Each group targets a different aspect of what makes a word easy or difficult to recall. Full definitions and sources are provided in \cref{sec:app:features}. Here, we describe the rationale behind each group.

\paragraph{Familiarity (11 features).}

These features capture how likely a learner is to have encountered a word before. We include logarithmic word frequency and contextual diversity from SUBTLEX-UK \citep{vanheuven-etal-2014-subtlexuk}, reported age of acquisition and percentage of annotators who knew the word \citep{kuperman-etal-2012-ageofacquisition}, CEFR level in the form of CEFR-J proficiency level, e.g., B2 for \texttt{cable}, and the EFLLex word-frequency profile across CEFR-graded books \citep{europarat-2011-common,negishi-etal-2013-progress,durlich-francois-2018-efllex}. Together, these features reflect both naturalistic exposure (how often the word appears in English media) and curricular sequencing (at which proficiency level the word is typically introduced).

\paragraph{Meaning (5 features).}

To approximate semantic complexity and concreteness, we use the $\ell_2$ norm of the English fastText embedding \citep{bojanowski-etal-2017-enriching}, mean hypernym depth and sense count from WordNet\footnote{For example, the word $\texttt{cable}$ has six senses, one of which has a hypernym path of depth $9$: cable $\subseteq$ conductor $\subseteq$ device $\subseteq$ instrumentality $\subseteq$ artifact $\subseteq$ whole $\subseteq$ object $\subseteq$ physical entity $\subseteq$ entity.} \citep{fellbaum-1998-wordnet}, the part-of-speech (POS) ratio from SUBTLEX-UK, and a binary flag indicating whether the KVL item required disambiguation, e.g., for the item \texttt{decode}, the German translation is \emph{etw entschlüsseln (nicht: decipher)}.

\paragraph{Surface (7 features).}

Surface features describe the orthographic form of the item as presented in the test. They include target- and source-word length in characters, number of syllables and letters per phoneme (an orthographic transparency proxy) of the target word, context-sentence length, and the first letter of both the English clue and the L1 translation. These features are relevant because the KVL task requires learners to reconstruct a spelling, making word length, letter cues, and phoneme density directly task-relevant \citep{james-burke-2000-phonological,beinborn-etal-2016-predicting}.

\paragraph{Transfer (1 feature).}

We compute the cosine similarity between character $n$-gram TF-IDF vectors of the English word and its L1 translation, see \cref{sec:app:charsim_details} for a detailed description. This feature captures orthographic overlap as a proxy for cognateness: High values indicate transparent cognates (e.g., English \texttt{fantasy} and Spanish \emph{fantas\'ia}), while the similarity is zero for all Chinese items due to the different script. Although this measure does not account for phonological similarity or regular sound correspondences, we find it to be highly informative for languages that share the Latin script with English.

\subsection{Model and baselines}
\label{sec:model}

For our purposes, we require a fast and interpretable model. We use CatBoost \citep{prokhorenkova-etal-2018-catboost}, a gradient-boosted decision-tree method that handles both numeric and categorical features. We train one model per L1 on the same set of 24 features (\cref{fig:graphical_abstract}, right), allowing each model to learn L1-specific interactions between features and difficulty. Hyperparameters are given in \cref{sec:app:hyperparameters}. Model training takes about ten seconds. To estimate variance, we train each model with 20 random seeds.

We compare against two baselines: a ridge regression model trained on the same features, which tests whether nonlinearity or feature interactions contribute beyond a linear combination, and the transformer-based approach of \citet{skidmore-etal-2025-transformer}, which fine-tunes a pretrained language model directly on KVL difficulty scores without handcrafted features.

\subsection{Evaluation}
\label{sec:evaluation}

We report the root-mean-square error (RMSE), in alignment with the BEA 2026 shared task \citep{felice-skidmore-2026-shared}, and Pearson's~$r$ on the test set. Our team's submission results (\emph{Boosted Cats -- HuDS lab}) are available at the \href{https://github.com/britishcouncil/bea2026st/blob/main/results/results_summary_test.md}{shared-task repository}. To test whether L1-specific models capture shared or distinct difficulty structures, we additionally evaluate each model on the test sets of the other two L1s.

For interpretability, we use tree-based SHAP values \citep{lundberg-lee-2017-unified,lundberg-etal-2018-consistent}, which attribute each prediction to individual feature contributions through Shapley values from cooperative game theory \citep{shapley-1953-value}. We aggregate absolute Shapley values by feature group to obtain their relative importance, analyzed in \cref{sec:mechanisms}.
We compute relative group importance as the sum of absolute Shapley values within each feature group, normalized to sum to one per item.


\section{Results}
\label{sec:results}

The same English word can be easy for one L1 group and hard for another, illustrated by the gold-label lexical difficulties that correlate across L1s but are far from identical, see \cref{sec:app:l1-correlation}. This observation motivates our approach to train L1-specific models, asking three questions. Are the model predictions accurate enough to warrant a study of how these predictions arise (\cref{sec:prediction-quality})? Which mechanisms do the models rely on, and how do these mechanisms differ by L1 (\cref{sec:mechanisms})? And finally, do models trained on one L1 generalize to another (\cref{sec:cross-l1_transfer})?

\subsection{Prediction quality}
\label{sec:prediction-quality}

\Cref{tab:prediction-results} compares our CatBoost models against two baselines: a linear (ridge) regression trained on the same features and the transformer-based approach of \citet{skidmore-etal-2025-transformer}. CatBoost achieves the lowest RMSE on all three L1s and the highest correlation in German and Chinese. The improvement over the linear baseline shows the importance of accounting for feature interactions and non-linear contributions to difficulty. CatBoost outperforms the transformer baseline, too, implying that well-chosen psycholinguistic features can perform on par with pretrained representations while being more easily interpretable.

\begin{table}[tbh]
\small
\centering
\setlength{\tabcolsep}{5pt}
\begin{tabular}{@{}l
    S[table-format=1.2]
    S[table-format=1.2]
    S[table-format=1.2]
    S[table-format=1.2]
    S[table-format=1.2]
    S[table-format=1.2]@{}}
\toprule
& \multicolumn{3}{c}{\textbf{RMSE} $\downarrow$}
& \multicolumn{3}{c}{\textbf{Pearson's} $r$ $\uparrow$} \\
\cmidrule(lr){2-4} \cmidrule(lr){5-7}
\textbf{Model}
  & {\textbf{ES}} & {\textbf{DE}} & {\textbf{CN}}
  & {\textbf{ES}} & {\textbf{DE}} & {\textbf{CN}} \\
\midrule
Transformer
  & 1.26 & 1.26 & 1.14
  & \textbf{0.77} & 0.77 & 0.75 \\
Linear regression
  & 1.30 & 1.20 & 1.07
  & 0.72 & 0.74 & 0.77 \\
CatBoost (ours)
  & \textbf{1.24} & \textbf{1.12} & \textbf{1.04}
  & 0.76 & \textbf{0.78} & \textbf{0.79} \\
\bottomrule
\end{tabular}
\caption{Test-set performance per L1. CatBoost results report the median over 20 random seeds; the $5$-$95\%$ range is $\leq 0.02$ for all entries. Best result per L1 in \textbf{bold}. The shared-task baseline fine-tunes a transformer on KVL difficulty scores
\citep{skidmore-etal-2025-transformer}.}
\label{tab:prediction-results}
\end{table}

\begin{figure}[tbhp]
    \centering
    \includegraphics[width=\linewidth]{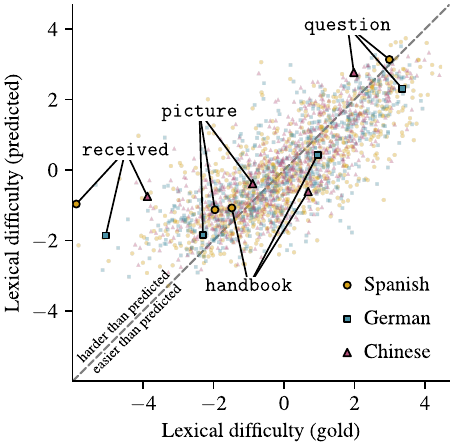}
    \caption{Predicted versus gold-label lexical difficulty. Each point is one L1-English vocabulary pair; the dashed diagonal marks a perfect prediction. Four exemplary items are highlighted: \texttt{question} (noun), \texttt{picture} (verb), \texttt{handbook} (noun), \texttt{received} (adjective). Seed variation of the prediction (5\textsuperscript{th}--95\textsuperscript{th} percentile) is small (median $0.1$, max.\ $< 0.4$).}
    \label{fig:pred_vs_gold}
\end{figure}

\Cref{fig:pred_vs_gold} shows predictions against the gold labels per L1. The predictions correlate well with the gold labels, as the bulk of items clusters near the diagonal (dashed gray). The hardest items, however, tend to be predicted as easier than they are. This bias is consistent across L1s, suggesting that the most difficult words involve challenges not captured entirely by our feature set, e.g., rare senses (\texttt{received} in the meaning of \enquote{generally accepted}). Other highlighted examples include \texttt{question}, which is easy and well-predicted for all L1s, although it is a cognate only in Spanish; \texttt{to~picture} and \texttt{handbook} show that our model captures the L1-dependent difficulty.

\begin{figure*}[tbh]
    \centering
    \includegraphics[width=\linewidth]{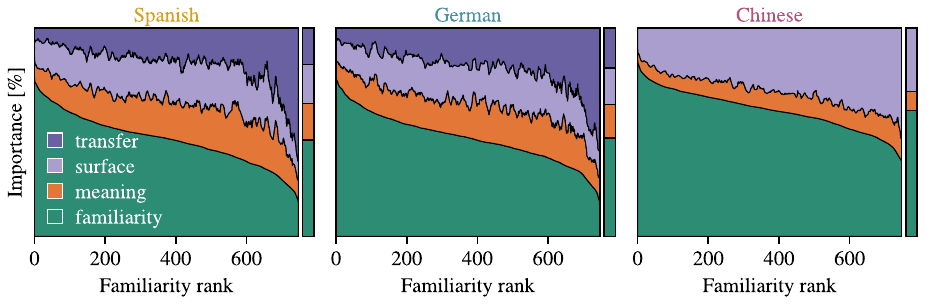}
    \caption{Per-item feature-group-importance shares, sorted by decreasing importance of familiarity (left to right). Median across 20 seeds, with a rolling average of window size 10. The bar at the right of each panel shows for each group the median across seeds of its mean importance share over all item. The \emph{form} group is split into \emph{transfer} (character similarity between the L1 and English) and \emph{surface} (features like word length and initial letters).}
    \label{fig:mechanism_stacked}
\end{figure*}

\subsection{Feature-group importance} 
\label{sec:mechanisms}

To understand the model predictions more intuitively, we thus group our 24 features into four feature groups---\emph{familiarity}, \emph{meaning}, \emph{surface}, and \emph{transfer}---and aggregate the absolute Shapley values per item for each group. The importance share of a feature group measures how strongly the features within this group contribute to predicting a specific item. We report the average importance of every feature and its correlation with the gold-label difficulty in \cref{tab:features}.


\paragraph{Transfer differentiates Spanish and German from Chinese learners.}

The most prominent cross-linguistic difference is the role of transfer. For Spanish and German speakers, a single feature---character-$n$-gram cosine similarity between the L1 and the English word---is the strongest predictor overall (mean $|\text{SHAP}|$ of 0.51 for Spanish and 0.52 for German; see \cref{fig:char_similarity} in \cref{sec:app:features}). For Chinese, this feature does not contribute at all because the Chinese writing system shares no characters with English.

This asymmetry is visible in \cref{fig:mechanism_stacked}, which shows the group importance of each item, sorted by decreasing importance of familiarity. Transfer fills the gap of less familiar words for Spanish and German. These are typically cognates, e.g., \texttt{handbook} and \emph{Handbuch}, which a German speaker can infer even without prior exposure to the English word.\footnote{Transfer beyond orthographic overlap exists but is not captured by character similarity. For example, the Chinese word \zh{手册} (\emph{sh\v{o}uc\`e}, literally \enquote{hand-book}) elicits transfer, too.} Transfer is most relevant for a distinct small subset of roughly 10--20\% of items relying least on familiarity (above rank 600).

\paragraph{Familiarity is strong across L1s.}

Familiarity accounts for the largest average feature-group importance for all three L1s, see the stacked bar summaries in \cref{fig:mechanism_stacked}. Within this group, the most important features differ: age of acquisition (AoA) and frequency for Spanish and German, versus the proficiency levels (EFLLex-level span and CEFR-J levels) for Chinese. This difference may reflect a variation in how English is acquired---naturalistic exposure for Spanish and German learners as opposed to potentially more structured, curriculum-driven learning for Chinese learners. Yet in all cases, familiar words are easier to recall.

\paragraph{Surface and meaning.}

Surface features are as important as meaning features (embedding norm, hypernym depth, and sense count) for Spanish and German, and substantially more important for Chinese, see \cref{fig:mechanism_stacked}. The importance of surface features is likely due to the KVL task format, which requires form recall rather than correct usage of a word. Spelling difficulty makes surface cues, e.g., word length and priming by the clue letter, directly task-relevant \citep{james-burke-2000-phonological,beinborn-etal-2016-predicting}. Chinese learners appear to rely on these cues more heavily.

\paragraph{Two routes to easiness.}

\begin{figure*}[tbh]
    \centering
    \includegraphics[width=\linewidth]{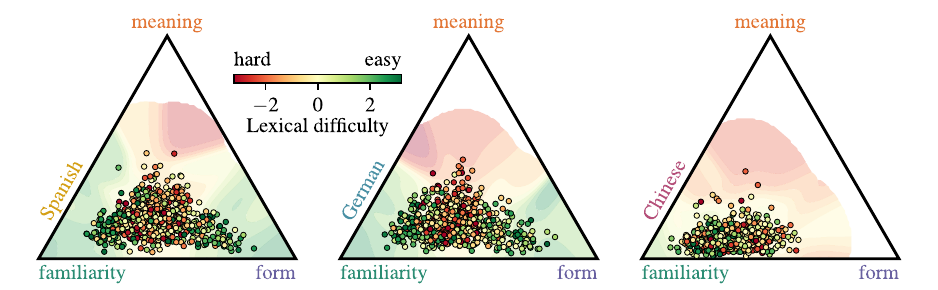}
    \caption{Each item projected onto a triangle according to the relative importance of three feature groups (familiarity, meaning, and form $=$ surface $\cup$ transfer). Color encodes the gold-label difficulty of an item (red: hard, green: easy). Background shading shows a regression of the difficulty surface; regions with insufficient data are masked.}
    \label{fig:mechanism_triangles}
\end{figure*}

\Cref{fig:mechanism_triangles} projects each item onto a triangle. Each corner represents that the respective feature group is uniquely important, while the middle corresponds to an equal mixture of importance. The color of each point encodes gold-label difficulty and the background shading is an estimate of the difficulty distribution in this triangle.\footnote{A Nadaraya--Watson kernel regression of the difficulty surface \citep{nadaraya-1964-estimating,watson-1964-smooth} with bandwidth selected by leave-one-out cross-validation \citep{hardle-1990-applied}.}

For Spanish and German, easy items (green) cluster in the two lower corners, familiarity and form. These clusters represent two routes through which a word is easy to a learner. A word is easy because it is frequent and early-acquired or because it closely resembles its L1 form. Some words are both common and have high orthographic similarity, which makes them easy as well (see \cref{sec:app:frequency_vs_charsim} for details), but they land in the middle region since the Shapley values for familiarity and transfer are both high. Difficult items (red) occupy the middle region in the triangle because neither cue is strong. 

For Chinese, the distribution in the triangle is instead tightly clustered along the lower-left familiarity-form edge\footnote{Chinese items show a somewhat lower dispersion in the feature-group simplex (Aitchison total variance: CN $0.30$, ES $0.33$, DE $0.35$ with bootstrap 90\% CIs < $0.05$).} and the background shading has no clear gradient. Chinese difficulty is shaped more uniformly by a combination of familiarity and form features.

\subsection{Cross-L1 transfer of models}
\label{sec:cross-l1_transfer}

\begin{figure}[tbhp]
    \centering
    \includegraphics[width=\linewidth]{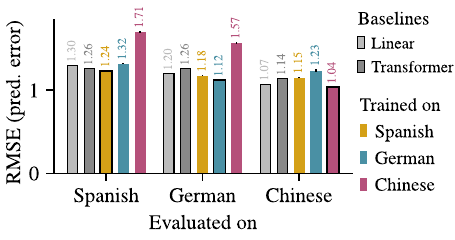}
    \caption{Cross-L1 evaluation. Each colored bar shows the RMSE prediction error (lower is better) of a CatBoost model trained on one L1 and evaluated on the L1 indicated on the horizontal axis. Thick black outlines mark within-L1 evaluation. Gray bars show the ridge linear baseline and the transformer baseline, each trained on the target L1. CatBoost results report the median over 20 random seeds with 5\textsuperscript{th}--95\textsuperscript{th} percentiles.}
    \label{fig:cross-L1-evaluation}
\end{figure}

If Spanish and German learners rely on similar feature groups, a model trained on one L1 should reasonably generalize to the other. We test this by evaluating each L1-specific CatBoost model on all three test sets, see \cref{fig:cross-L1-evaluation}.

Each model performs best on its own L1, but the degradation when transferring between Spanish and German is minor. The Spanish-trained model achieves RMSE~1.18 on German (vs.\ 1.12 within-German), and the German-trained model reaches RMSE~1.32 on Spanish (vs.\ 1.24 within-Spanish)---both on par with the linear baselines (RMSE 1.30 for Spanish, 1.20 for German) and the transformer baselines (RMSE 1.26 for Spanish, 1.26 for German) trained directly on the target L1. A model trained on Spanish or German thus captures meaningful mechanisms that transfer to the other, without any access to target-L1 training data.

Transfer involving Chinese is markedly worse in both directions. Chinese-trained models yield RMSE~1.71 on Spanish and 1.57 on German, well above the baselines. Spanish- and German-trained models evaluated on Chinese (RMSE~1.15 and 1.23) degrade less severely, which aligns with the observation that the Spanish and German models have learned all the features relevant to predicting difficulty for Chinese learners, although they still fall short of the within-Chinese model (RMSE~1.04). The Chinese-trained model, not having learned to predict items based on transfer, cannot generalize to Spanish or German.

These cross-L1 results mirror the feature-group analysis: Spanish and German models share a difficulty landscape, while the prediction for Chinese learners relies on different cues. Practically, this observation suggests that models trained on a given L1, say Spanish, could provide useful estimates for typologically related languages such as Portuguese or Italian, even without the dedicated training data.


\section{Discussion}
\label{sec:discussion}

Our finding that word familiarity exerts the strongest impact on the KVL difficulty scores across all L1 groups aligns with prior research identifying word frequency and age of acquisition as robust predictors of lexical processing \citep{rott-1999-effect,ellis-2002-frequency,kuperman-etal-2012-ageofacquisition,north-etal-2023-lexical}. Our analysis adds a quantitative decomposition that reveals \emph{how} this shared component interacts with L1-specific factors. For Spanish and German, transfer provides a second route to easiness: Words can be easy because they are familiar or because they are orthographic cognates. This two-route structure aligns with \citet{urdaniz-skoufaki-2022-spanish}, who find frequency and cognateness to be strong, interacting predictors for Spanish learners of English, and with \citet{degroot-keijzer-2000-what}, who show that cognates are easier to remember. We observe the same pattern for German learners. For Chinese learners, difficulty is predicted more uniformly by surface features and familiarity. The cognitive strategies during form recall thus appear to differ qualitatively, depending on the typological relationship between the L1 and L2.

Our character similarity feature captures character overlap as a proxy for perceived similarity \citep{ringbom-1987-role}, which need not coincide with etymological relatedness. 
The strength of this feature in our models suggests that even coarse character overlap is a powerful retrieval cue when L1 and L2 share a script. Although Chinese has no orthographic overlap, cross-linguistic transfer may still be present in the form of loanwords (e.g., \zh{雷达} \emph{l\'eid\'a} from English radar), calques (brainwash from the Chinese \zh{洗脑} \emph{x\v{\i}n\v{a}o}), and parallel compositions (\zh{星光} \emph{x\={\i}nggu\=ang} happening to map to \enquote{starlight}) or morphology (\zh{地}
\emph{de} marking adverbs like the English suffix \enquote{-ly}).

The task format affects which aspects of word knowledge are tested \citep{laufer-goldstein-2004-testing,culligan-2015-comparison}. In our case, meaning features contribute little across all L1s, likely because the KVL task tests spelling rather than an understanding of meaning. A task requiring contextual usage would likely involve more meaning-related features.

A caveat concerns interpreting Shapley values as cognitively plausible. Our models approximate human test responses, and their strong held-out performance suggests that the features are plausible factors in the learner behavior that underlies the test responses. Where the model fails, however, the relevant factors likely lie outside our feature set, e.g., individual learner differences, contextual effects of the context sentence, or interference from false friends. We discuss the relation of lexical features beyond our predictor set with the gold-label difficulty and model-prediction errors in \cref{sec:app:redundant_features}.

Our findings also have practical implications. Spanish and German learners could benefit from curricula rich in cognates in order to efficiently build an extensive vocabulary \citep{nation-2000-learning}. Furthermore, a model trained on data from one L1 could serve in predicting lexical difficulty for typologically related, lower-resource languages without dedicated data. For Chinese learners of English (and potentially speakers of languages not related to the target language), a curriculum based on frequency and proficiency levels is likely more effective. Since our model can generalize to unseen words, we provide predictions of lexical difficulty beyond the KVL dataset, on the entire SUBTLEX-UK vocabulary for which we could compute the features. Our model predictions can be explored \href{https://huds.pages.gwdg.de/projects/bea2026}{interactively}.


\section{Conclusions}
\label{sec:conclusion}

We have investigated what makes an English word difficult for second-language learners by training interpretable models on lexical-difficulty scores from Spanish, German, and Chinese L1 speakers. By decomposing the predictions into feature groups, we identify two qualitatively different profiles. Spanish and German learners benefit from two routes to easiness---familiarity and orthographic transfer---while Chinese learners rely on a less structured combination of familiarity and surface features in the absence of orthographic overlap. The models of learners with Spanish and German as L1 generalize well to each other, indicating that the learned difficulty functions reflect shared cognitive strategies.

For future research, the strong cross-L1 transfer between the Spanish- and German-trained models suggests that models trained on a high-resource L1 could provide useful difficulty estimates for typologically related languages (e.g., from Spanish to Portuguese, Catalan, or Italian; potentially also from Chinese to other languages not related to English like Japanese), even without the dedicated training data. 
Another promising direction is to investigate whether the identified mechanisms are symmetric, i.e., whether the same profile of feature-group importance applies for English speakers that learn Spanish, German, or Chinese. Third, incorporating phonological similarity and including calques and morphological similarity could capture part of the variance in lexical difficulty that our current feature set does not explain.

\section*{Limitations}

\paragraph{Language coverage.} The KVL data covers a limited selection of four high-resource languages, three of which are Indo-European. The extent to which the patterns we identify generalize to learners with other language backgrounds remains an open question.

\paragraph{Task specificity.} The KVL dataset tests productive form recall only. Our analysis therefore pertains to this aspect of vocabulary knowledge and could be extended to other test formats.

\paragraph{Measuring transfer.} We operationalize cross-linguistic transfer through character-level overlap, which does not capture phonological similarity, regular sound correspondences, and semantic transfer (e.g., calques).

\paragraph{Interpretation of feature importance.} Feature-group importance describes the structure of model predictions and cannot measure a learner's cognitive processes directly. Inferences about human cognition thus remain indirect.

\paragraph{Curricular effects.} Lexical difficulty is not only influenced by word properties but also the curriculum order, which is captured by our CEFR-proficiency-level features. Beyond the intrinsic difficulty of a word, the KVL difficulty we model therefore partially encodes the status quo in instruction regarding where a word happens to be placed in educational curricula.

\section*{Ethical considerations}

This work is intended as fundamental research into L2 vocabulary difficulty. While our results may inform applications and further research, it should not be interpreted as prescriptive guidance for teaching in practice.

Our computational experiments use light-weight models that entail no substantial environmental impact.

\section*{Acknowledgments}

We thank the reviewers for their thoughtful and detailed feedback. This research is partially supported by the zukunft.niedersachsen program of the VolkswagenStiftung (L.B., Z.H.) and by a VENI grant (Vl.Veni.211C.039) from the Nederlandse Organisatie voor Wetenschappelijk Onderzoek (NWO) (L.B., A.H.).

\section*{Author contributions}

\textbf{J.M.M.:} Conceptualization, Analysis, Software, Literature review, Visualization, Writing---original draft, Writing---review \& editing.
\textbf{Z.H.:} Analysis and Software (feature analysis for App.\ G, KVL extension), Literature review, Visualization (App.\ G), Writing---original draft (App.\ E, F, and G; parts of Introduction and Related Work), Writing---review \& editing.
\textbf{A.H.:} Literature review, Writing---original draft (parts of Related Work).
\textbf{L.B.:} Supervision, Conceptualization, Analysis, Literature review, Writing---review \& editing.

\bibliography{refs}

\appendix
\crefalias{section}{appendix}

\section{Demo}
\label{sec:app:demo}

\begin{figure*}[tbh]
    \centering
    \includegraphics[width=\linewidth]{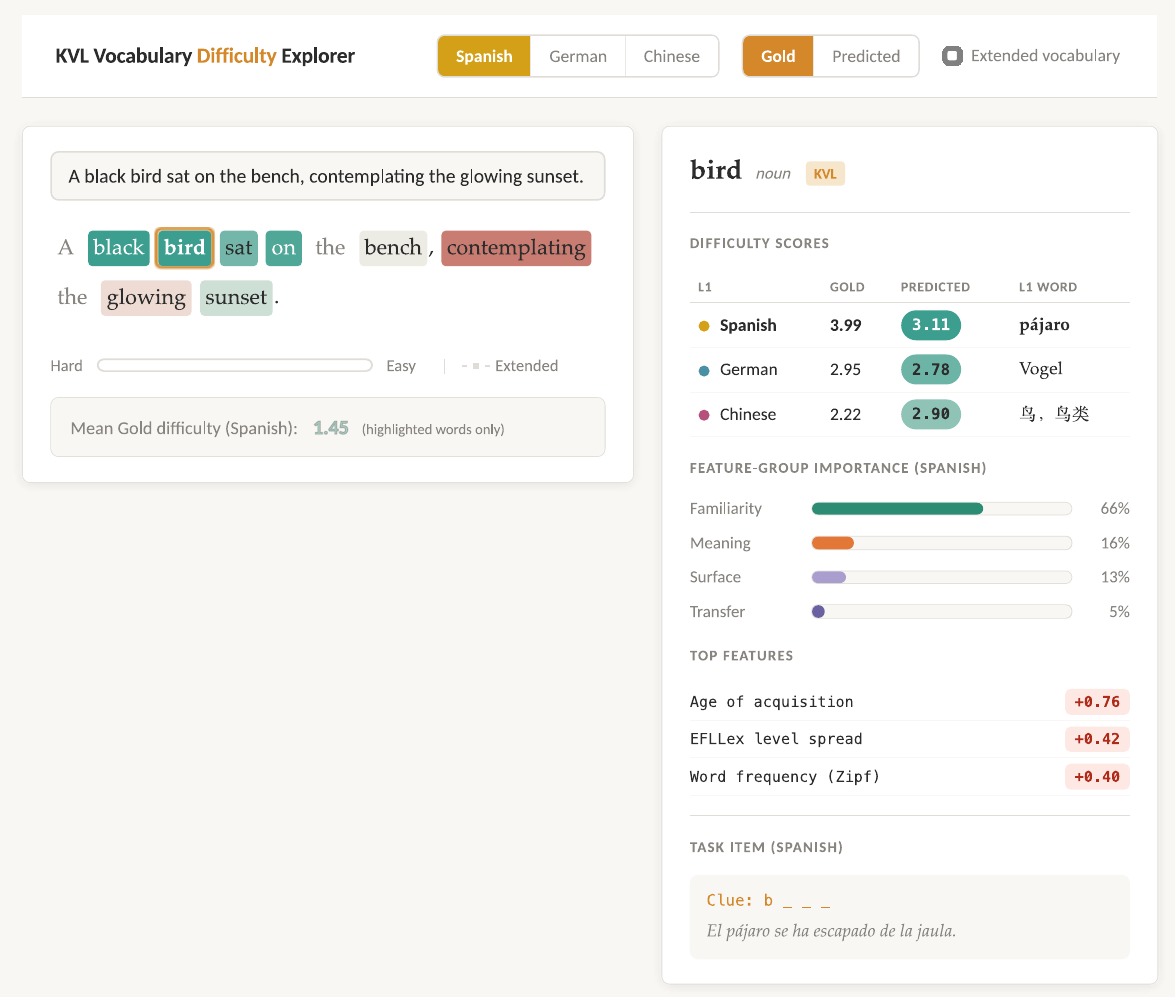}
    \caption{Screenshot of our \href{https://huds.pages.gwdg.de/projects/bea2026}{interactive demo}. Words of the input text are highlighted according to their lexical difficulty. Clicking on a word opens a panel that shows the gold-label and predicted difficulty as well as the feature-group importance. The L1-background can be switched.}
    \label{fig:demo}
\end{figure*}

We provide an \href{https://huds.pages.gwdg.de/projects/bea2026}{interactive demo} that allows users to explore the KVL dataset and our model predictions, see \cref{fig:demo}. Given an input text, words are highlighted according to their predicted or gold-label lexical difficulty. Selecting a word opens a panel showing its lexical difficulty, feature-group importance, top individual features, clue letter and context clue.

To map a word form in the input text to KVL lemmas, we precompute a lookup table of inflected forms using the \textsc{PyInflect} library. Because the demo is implemented as a static web application, it does not perform syntactic parsing on the fly. Instead, the most frequent part of speech is shown by default, with the option to select any alternatives. The interface also allows users to switch the language background and optionally include our extended vocabulary (see \cref{sec:app:KVL_extension}).  

\section{KVL extension}
\label{sec:app:KVL_extension}

To support lexical-difficulty predictions beyond the original KVL inventory, we construct an extended vocabulary in the KVL task format derived from \textsc{Wiktionary} data.

\paragraph{Source and filtering.}

We begin with Wiktionary JSONL extracts produced by
\textsc{Wiktextract} \citep{ylonen-2022-wiktextract}. From these, we retain only English headwords ($\text{\texttt{lang\_code}} = \text{en}$) that appear in SUBTLEX-UK \citep{vanheuven-etal-2014-subtlexuk}. This filtering restricts the extension to commonly used words. We further discard entries that are empty, single-character, or all-uppercase (e.g., acronyms). Language-specific filters remove entries that primarily encode inflectional variants, spelling alternatives, or other metalinguistic notes.

\paragraph{Normalization.}
The retained entries are converted into a compact, standardized representation. From each Wiktionary entry, we extract only the information necessary for the KVL format: the English target word and part of speech, a masked spelling clue, an L1 translation, and an optional L1 context sentence. Duplicate English headwords are removed by keeping the first entry with a valid part-of-speech annotation.

\paragraph{L1-specific processing.}

For German and Spanish, the L1 source word is taken from the first available translation or, if unavailable, from the shortest gloss phrase. For Chinese, if no translation is available, the L1 source word is extracted from gloss blocks and normalized, including conversion to simplified characters. Context sentences are derived from example translations.

\paragraph{Excluding overlap.}
To ensure that the extended vocabulary contains no duplicates with the original KVL data, we exclude all English target words that occur in the KVL training, development, or test splits for the corresponding L1. The final output consists of three L1-specific CSV files that follow the original KVL schema and add 7{,}434 items for Chinese, 4{,}325 for German, and 6{,}606 for Spanish.

\section{Features} 
\label{sec:app:features}

\begin{table*}[t]
\small
\centering
\setlength{\tabcolsep}{3pt}
\begin{tabular}{@{}clp{5.3cm}l
    ccc
    rrr@{}}
\toprule
& & & &
  \multicolumn{3}{c}{\textbf{Mean $|\text{SHAP}|$}} &
  \multicolumn{3}{c}{\textbf{$\rho$ (value-gold)}} \\
\cmidrule(lr){5-7} \cmidrule(lr){8-10}
& \textbf{Feature} & \textbf{Definition} & \textbf{Source}
  & \textbf{ES} & \textbf{DE} & \textbf{CN}
  & \textbf{ES} & \textbf{DE} & \textbf{CN} \\
\midrule
\multirow{11}{*}{\rotatebox[origin=c]{90}{Familiarity}}
 & Log frequency
 & Log-scaled word frequency
 & SUBTLEX-UK$^1$
 & \textbf{0.26} & \textbf{0.20} & \textbf{0.19}
 & $.40$ & $.43$ & $.60$ \\
 & Contextual diversity
 & Distinct film contexts
 & SUBTLEX-UK$^1$
 & 0.05 & 0.06 & \textbf{0.16}
 & $.37$ & $.39$ & $.58$ \\
 & Age of acquisition
 & Mean AoA rating
 & AoA$^2$
 & \textbf{0.27} & \textbf{0.27} & \textbf{0.20}
 & $-.44$ & $-.46$ & $-.55$ \\
 & Percent known
 & \% of raters knowing the word
 & AoA$^2$
 & 0.02 & 0.03 & 0.02
 & $.17$ & $.18$ & $.22$ \\
 & CEFR-J level
 & Vocabulary level (A1\,=\,1 \ldots\ C2\,=\,6)
 & CEFR-J$^3$
 & 0.15 & 0.17 & \textbf{0.23}
 & $-.55$ & $-.57$ & $-.66$ \\
 & EFLLex-level span
 & CEFR levels with non-zero frequency
 & EFLLex$^4$
 & \textbf{0.25} & \textbf{0.22} & \textbf{0.31}
 & $.43$ & $.45$ & $.64$ \\
 & EFLLex A1
 & Learner-corpus token share at A1
 & EFLLex$^4$
 & 0.06 & 0.04 & 0.02
 & $.40$ & $.42$ & $.49$ \\
 & EFLLex A2
 & \quad\textit{''} at A2
 & \textit{''}
 & 0.13 & 0.10 & 0.05
 & $.40$ & $.48$ & $.52$ \\
 & EFLLex B1
 & \quad\textit{''} at B1
 & \textit{''}
 & 0.05 & 0.06 & 0.03
 & $.30$ & $.36$ & $.46$ \\
 & EFLLex B2
 & \quad\textit{''} at B2
 & \textit{''}
 & 0.03 & 0.05 & 0.08
 & $.16$ & $.18$ & $.34$ \\
 & EFLLex C1
 & \quad\textit{''} at C1
 & \textit{''}
 & 0.03 & 0.03 & 0.03
 & $.05$ & $.05$ & $.18$ \\
\midrule
\multirow{5}{*}{\rotatebox[origin=c]{90}{Meaning}}
 & Embedding norm
 & fastText embedding $\ell_2$ norm
 & fastText$^5$
 & \textbf{0.23} & \textbf{0.18} & 0.08
 & $.32$ & $.33$ & $.33$ \\
 & Hypernym depth
 & Mean synset depth in hypernym tree
 & WordNet$^6$
 & 0.13 & 0.11 & 0.03
 & $.15$ & $.18$ & $.05$ \\
 & Sense count
 & Number of synsets for word\,+\,POS
 & WordNet$^6$
 & 0.07 & 0.04 & 0.03
 & $.12$ & $.14$ & $.24$ \\
 & POS dominance ratio
 & Token share of dominant POS
 & SUBTLEX-UK$^1$
 & 0.04 & 0.03 & 0.05
 & $.00$ & $-.01$ & $-.07$ \\
 & L1 confusor flag
 & L1 source had disambiguation annotation
 & KVL$^7$ (derived)
 & 0.02 & 0.05 & 0.01
 & $-.08$ & $-.12$ & $-.09$ \\
\midrule
\multirow{7}{*}{\rotatebox[origin=c]{90}{Surface}}
 & Target word length
 & Character length of English word
 & KVL$^7$ (derived)
 & 0.04 & 0.04 & 0.10
 & $-.33$ & $-.35$ & $-.43$ \\
 & Source word length
 & Character length of L1 lemma
 & KVL$^7$ (derived)
 & 0.04 & 0.05 & 0.15
 & $-.23$ & $-.36$ & $-.33$ \\
 & Syllable count
 & Estimated syllables of English word
 & KVL$^7$ (derived)
 & 0.05 & 0.05 & 0.05
 & $-.27$ & $-.32$ & $-.37$ \\
 & Letters per phoneme
 & Orthographic transparency proxy
 & AoA$^2$
 & 0.04 & 0.04 & 0.02
 & $.08$ & $.01$ & $.10$ \\
 & Context sent.\ length
 & Character length of L1 context sentence
 & KVL$^7$ (derived)
 & 0.10 & 0.07 & 0.11
 & $-.21$ & $-.24$ & $-.30$ \\
 & Clue letter
 & First letter of the spelling clue
 & KVL$^7$
 & 0.18 & 0.12 & 0.16
 & {---} & {---} & {---} \\
 & L1 initial letter
 & First letter of the L1 lemma
 & KVL$^7$ (derived)
 & 0.06 & 0.07 & 0.07
 & {---} & {---} & {---} \\
\midrule
\multirow{1}{*}{\rotatebox[origin=c]{90}{Transfer}}
 & & & & & & & & & \\
 & Character similarity
 & Char.\ $n$-gram TF-IDF cosine, EN vs.\ L1
 & KVL$^7$ (derived)
 & \textbf{0.51} & \textbf{0.52} & 0.00
 & $.10$ & $.25$ & {---} \\
 & & & & & & & & & \\
\bottomrule
\end{tabular}
\vspace{4pt}
\raggedright
\footnotesize
$^1$\citet{vanheuven-etal-2014-subtlexuk};\quad
$^2$\citet{kuperman-etal-2012-ageofacquisition};\quad
$^3$\citet{negishi-etal-2013-progress};\quad
$^4$\citet{durlich-francois-2018-efllex};\\
$^5$\citet{bojanowski-etal-2017-enriching};\quad
$^6$\citet{miller-1995-wordnet, fellbaum-1998-wordnet};\quad
$^7$\citet{schmitt-etal-2021-introducing,skidmore-etal-2025-transformer}
\caption{Mean $|\text{SHAP}|$ importance per feature and L1, averaged over test-set items
across 20 random seeds ($5$--$95\%$ range $<0.02$ for all features and L1s),
and Spearman's $\rho$ between feature value and gold lexical difficulty.
Features are grouped by category.
The five highest $|\text{SHAP}|$ values per L1 column are set in \textbf{bold}.}
\label{tab:features}
\end{table*}

\Cref{tab:features} lists the 24 features that we use for our models, grouped by mechanism. For each feature, we report the median absolute Shapley value per L1, computed by averaging absolute Shapley values over test‑set items and then taking the median across 20 random seeds. We additionally report Spearman's rank correlation between the feature value and gold-label difficulty over test-set items.

Intuitively, the Shapley values quantify how important a feature is on average for the model predictions. Spearman's rank correlations show how monotonic the relation is between the feature value and lexical difficulty. Particularly interesting are those features for which the two measures do not align. For instance, character similarity has a weak correlation with difficulty for Spanish ($\rho = .10$) and German ($\rho = .25$), yet it is the feature with the highest Shapley-value importance for these two languages. This is in part because transparent cognates are relatively rare, and similarity is high only for this subset of transparent cognates. We discuss this in \cref{sec:app:charsim_details}. Moreover, the character similarity, as we show in our main analysis \cref{sec:mechanisms}, interacts with familiarity such that this transfer feature tends to be important when familiarity is not available.
Conversely, the CEFR-J level has the highest Spearman's $|\rho|$ across all L1s (up to $-.66$ for Chinese) but is only moderately important because it is collinear with other familiarity features, e.g., frequency and EFLLex-level span.

We briefly define features that are not self-explanatory:

\paragraph{Logarithmic frequency.}
Log-transformed word frequency on the Zipf scale \citep{vanheuven-etal-2014-subtlexuk}
\begin{equation}
    f_{\text{Zipf}} = \log_{10}\!\bigl(\text{fpmw}\bigr) + 3
\end{equation}
which modifies the standardized measure of frequency per million words (fpmw) by a logarithmic transformation and adding $3$ to keep the measure positive even for low-frequency words, making it more intuitive. We assign a value below the minimal frequency threshold, $f_{\min} - 0.5$, to words missing from the SUBTLEX-UK list.

\paragraph{Contextual diversity.}
The proportion of film or television programs in SUBTLEX-UK containing at least one occurrence of the word, reflecting the breadth of contexts in which a learner might encounter it \citep{vanheuven-etal-2014-subtlexuk}. We expect that words that occur only in certain domains are less familiar (and thus more difficult to learn) relative to words of the same frequency that occur across many domains. This feature turns out to be especially relevant for Chinese-speaking learners.

\paragraph{EFLLex level span.}
The number of Common European Framework of Reference for languages (CEFR) levels \citep{europarat-2011-common}, A1, A2, B1, B2, C1, at which the word has non-zero frequency in the EFLLex learner corpus \citep{durlich-francois-2018-efllex}. A~span of 5 means the word appears at every level from A1 to C1 and is thus broadly familiar, whereas a span of 1 means it is confined to a single proficiency band.

\paragraph{Embedding norm.}
The $\ell_2$ norm of the 300-dimensional English fastText embedding $\mathbf{e}$ \citep{bojanowski-etal-2017-enriching},
\begin{equation}
    \lVert \mathbf{e} \rVert_2 = \sqrt{\textstyle\sum_{i=1}^{300} e_i^2}\,.
\end{equation}
Since function words and polysemous words tend to have smaller norms because their embeddings are pulled towards the origin of the space by diverse contexts, concrete and semantically specific words tend to have larger norms \citep{oyama-etal-2023-norm}.

\paragraph{Character similarity.}
We represent the L1 word and its English translation as character $n$-gram TF-IDF vectors ($n\in\{2, 3, 4\}$, sublinear TF) and compute their cosine similarity. For Chinese, all values are approximately zero because there is no character overlap between Latin and Chinese script. We give a more detailed definition in \cref{sec:app:charsim_details}.

\paragraph{POS dominance ratio.}
The proportion of SUBTLEX-UK tokens for a lemma that carry its most frequent part of speech (POS), derived from the POS frequency in SUBTLEX-UK \citep{vanheuven-etal-2014-subtlexuk}. Values near 1 indicate unambiguous POS. Lower values indicate that the word is frequently used as multiple parts of speech (e.g., \texttt{light} as noun, verb, and adjective).

\paragraph{Letters per phoneme.}
The ratio of character length to phoneme count, derived from the number of phonemes per word in the \citet{kuperman-etal-2012-ageofacquisition} dataset. This serves as a proxy for orthographic transparency: Words with many letters per phoneme have less predictable spellings, e.g., \texttt{thought} \textipa{/T\textopeno:t/} (7 letters / 3 phonemes = 2.33).

\paragraph{Clue letter and L1 initial letter.}
These categorical features encode the first letter of the English spelling clue and the first letter of the L1 lemma, respectively. They capture letter-level priming effects: certain initial letters may be associated with easier or harder vocabulary on average, and letter overlap between the clue and the L1 form may facilitate recall. Spearman correlations are not reported for categorical features. Interestingly, the L1-first-letter priming is equally important for Chinese as for Spanish and German ($0.06$--$0.07$) as the model learns a baseline for every Chinese word-initial character.

\begin{figure}[tbhp]
    \centering
    \includegraphics[width=\linewidth]{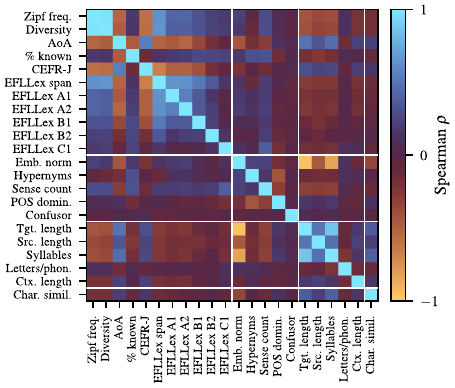}
    \caption{Pairwise Spearman correlation between numeric features (Spanish; the pattern is similar for German and Chinese except for L1-dependent features such as source word length and character similarity). Several pairs are substantially correlated. White lines indicate feature groups.}
    \label{fig:feature_corr}
\end{figure}

\Cref{fig:feature_corr} shows pairwise correlations between the numeric features from \cref{tab:features}. Some features are strongly correlated, particularly the familiarity features (e.g., logarithmic frequency, contextual diversity, and EFLLex level span) and the surface features (target word length, syllable count, and source word length). This collinearity means that individual SHAP values within a group partly reflect shared variance, which motivates our decision to group the features for the main analysis.

Additional features were evaluated during development but excluded from the final model after unstructured ablation showed no consistent improvement. These include: character-level surprisal, normalized Levenshtein edit distance, longest common subsequence ratios, EFLLex entropy and mean level, WordNet synonym count and derivational family size, POS competition flag, and Spanish-specific CORPES frequencies and anglicism indicators. We provide these in \cref{tab:ablated_features} for completeness. These features turn out to be too redundant to yield higher performance. For example, edit distance captures the same signal as character similarity but is less informative in this context. We give an exemplary analysis in \cref{sec:app:redundant_features}.

\begin{table*}[t]
\small
\centering
\setlength{\tabcolsep}{4pt}
\begin{tabular}{@{}ll}
\toprule
\textbf{Feature} & \textbf{Definition} \\
\midrule
Character surprisal
  & Mean $-\!\log(p(\text{char}))$ under training-set unigram distribution \\
Edit distance (norm.)
  & Levenshtein distance between EN and L1 word, normalized by $\sqrt{\text{len}_{\text{EN}}} \cdot \sqrt{\text{len}_{\text{L1}}}$ \\
LCS ratio (EN)
  & Longest common subsequence length / target word length \\
LCS ratio (L1)
  & Longest common subsequence length / source word length \\
EFLLex entropy
  & Normalized Shannon entropy of the CEFR-level frequency distribution \\
EFLLex mean level
  & Frequency-weighted mean CEFR level \\
WN synonym count
  & Number of unique lemma names across all synsets \\
WN derivational family
  & Number of derivationally related forms in WordNet \\
WN POS count
  & Number of distinct parts of speech in WordNet \\
POS competition
  & 1 if the tested POS differs from the dominant SUBTLEX-UK POS \\
Exact match
  & 1 if EN word = L1 lemma after normalization \\
L1 second letter
  & Second letter of the L1 lemma \\
CORPES frequency (ES)
  & Log-transformed normalized frequency in the Spanish CORPES corpus \\
CORPES dispersion (ES)
  & Log-transformed dispersion across CORPES subcorpora \\
Anglicism frequency (ES)
  & CORPES frequency of the English word looked up in Spanish text \\
Anglicism flag (ES)
  & 1 if the L1 lemma appears in a Wiktionary-derived anglicism list \\
\bottomrule
\end{tabular}
\caption{Additional features evaluated on the development set but excluded from the final model after ablation showed no consistent improvement of model performance. CORPES refers to \citet{realacademiaespanola-2025-corpus}.}
\label{tab:ablated_features}
\end{table*}
\begin{figure}[tbhp]
    \centering
    \includegraphics[width=\linewidth]{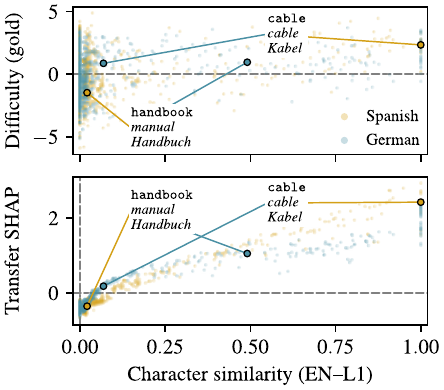}
    \caption{Character similarity between English words and their L1 translations (Spanish, German) versus gold-label difficulty (top) and Shapley-value contribution of this feature (bottom). Two words are highlighted: \texttt{cable} and \texttt{handbook}.}
    \label{fig:char_similarity}
\end{figure}

\section{Character similarity}
\label{sec:app:charsim_details}

To approximate cross-linguistic form similarity between an English word and its L1 translation, we compare their character-$n$-gram overlap. This feature is motivated by the intuition that words sharing many letter sequences (e.g., \texttt{fantasy} and \emph{fantas\'ia}) can be readily recognized as cognates.

\paragraph{From binary overlap to weighted similarity.}

In a naive approach, we could represent each word as a binary vector over character $n$-grams, where each entry states whether a given $n$-gram occurs in the word, and then compute the cosine similarity between two vectors. This approach would treat all $n$-grams the same, yet some $n$-grams might be more informative than others.

To account for the informativeness of an $n$-gram, we weight it using \emph{term-frequency--inverse-document-frequency} (TF-IDF). In this context, words correspond to \enquote{documents} and $n$-grams with $n \in \{2,3,4\}$ to \enquote{terms}. The TF-IDF weight of an $n$-gram $t$ in a word $w$ is implemented as
\begin{equation}
    \text{tf-idf}(t, w) = \big(1 + \log(\text{tf}(t, w))\big) \log\!\bigg(\frac{N}{\text{df}(t)}\bigg)\,,
\end{equation}
where $\text{tf}(t,w)$ is the frequency of $t$ in $w$, $\text{df}(t)$ is the number of words in the training data containing $t$, and $N$ is the total number of word forms in the union of English and L1 word forms in the training data. We use sublinear term-frequency scaling, i.e., $1 + \log(\text{tf}(t, w))$ instead of $\text{tf}(t, w)$, so that repeated occurrences are weighted less than linearly.

\paragraph{Cosine similarity.}

We can thus represent each word as a TF-IDF-weighted vector over character $n$-grams. Cross-linguistic form similarity is then measured as the cosine similarity between the English word $w_{\text{en}}$ and its L1 translation $w_{\text{L1}}$:
\begin{equation}
    \text{sim}(w_{\text{en}}, w_{\text{L1}}) =
    \frac{
        \mathbf{x}_{w_{\text{en}}} \cdot \mathbf{x}_{w_{\text{L1}}}
        }{
        \lVert \mathbf{x}_{w_{\text{en}}} \rVert \lVert \mathbf{x}_{w_{\text{L1}}} \rVert}.
\end{equation}
The TF--IDF vectors are $\ell_2$-normalized, so cosine similarity corresponds to their dot product.

\paragraph{Empirical behavior.}

\Cref{fig:char_similarity} illustrates how this character similarity relates to lexical difficulty and to its feature importance in Spanish- and German-L1 model predictions. Most items have near-zero similarity, with some items spread across the range up to perfect matches at 1. Higher similarity is associated with lower difficulty, cf.\ \cref{tab:features}. The Shapley values show that high similarity contributes positively to predictions above a small threshold at approximately 0.05--0.10. Below this threshold, negligible similarity is a cue for higher difficulty. This nonlinear distribution explains why character similarity has high average feature importance despite only a moderate correlation with difficulty.

\begin{figure*}[tbhp]
    \centering
    \includegraphics[width=\linewidth]{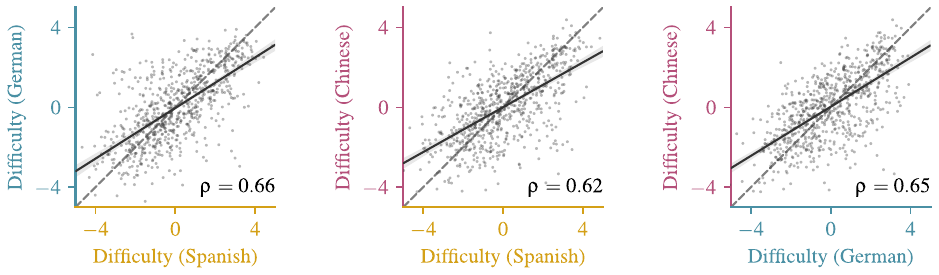}
    \caption{Pairwise correlation of gold-label lexical difficulty across L1s. Each point is one English word tested in both languages. Pearson's $r$ is shown per pair. The shared variance reflects universal difficulty factors and the unexplained variance motivates L1-specific modeling.}
    \label{fig:L1-correlation}
\end{figure*}

\section{Hyperparameters}
\label{sec:app:hyperparameters}

Relevant hyperparameters are summarized in \cref{tab:model_config}. The hyperparameters were tuned on the Spanish development set. Reported results are based on 20 random seeds.

\begin{table}[ht]
\small
\centering
\setlength{\tabcolsep}{5pt}
\begin{tabular}{@{}lll@{}}
\toprule
\textbf{Component} & \textbf{Parameter} & \textbf{Value} \\
\midrule
\multirow{6}{*}{CatBoost}
 & Loss function      & RMSE \\
 & Tree depth         & 7 \\
 & Learning rate      & 0.017 \\
 & Iterations         & 2{,}400 \\
 & $\ell_2$ leaf regularization & 0.8 \\
\midrule
\multirow{3}{*}{Evaluation}
 & Eval.\ seeds      & 20 (1, 8, 15, \ldots, 134) \\
 & Submission seeds  & 3 (42, 142, 242) \\
 & Metrics           & RMSE (primary) \\
 & & Pearson's $r$ \\
\midrule
Baseline & Model & Ridge regression \\
& $\ell_2$ regularization & 1 \\
\bottomrule
\end{tabular}
\caption{CatBoost model configuration, evaluation setup, and baseline.} 
\label{tab:model_config}
\end{table}

\section{Correlation of difficulties across L1s}
\label{sec:app:l1-correlation}

\Cref{fig:L1-correlation} shows pairwise correlations of gold-label difficulty across L1s. All language pairs are positively correlated, quantifying the shared component of lexical difficulty: Words that are hard for one L1 group tend to be hard for others. However, the correlations are far from perfect, indicating that L1-specific factors account for a substantial portion of the variance.

\section{Relationship of frequency and transfer}
\label{sec:app:frequency_vs_charsim}

\begin{figure*}[tbhp]
    \centering
    \includegraphics[width=\linewidth]{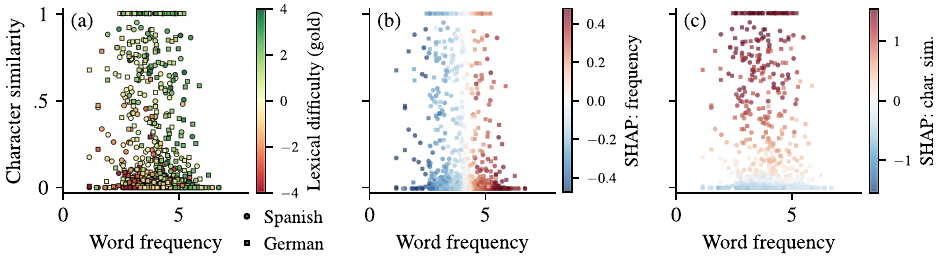}
    \caption{Character similarity by word frequency for Spanish and German, colored by (a) lexical difficulty and the feature importances of (b) word frequency and (c) character similarity.}
    \label{fig:frequency_vs_charsim}
\end{figure*}

In \cref{fig:frequency_vs_charsim}, panel (a) shows lexical difficulty in the plane of character similarity and word frequency. We observe that the easiest words are frequent or cognate. Those words that are both cognate and frequent (items in the top right) shape the model predictions equally through both features, as the Shapley values in panels (b) and (c) indicate. The thresholds for the Shapley values that the model learns appear largely independent for these two features: a vertical divide.

\section{Feature analysis}
\label{sec:app:redundant_features}

In this section, we examine several features that we evaluated during development but excluded from the final model, cf.\ \cref{tab:ablated_features}. The goal is to show how these features are related to lexical difficulty and why they may be redundant.

\paragraph{POS competition.}
Some English word forms occur frequently with one part of speech while another is less common, e.g., \texttt{received} as an adjective meaning \enquote{accepted} or \texttt{bar} as verb rather than a noun. We define a binary POS-competition flag, i.e., whether the POS of the target word differs from the most frequent POS. 

\begin{figure}[tbhp]
    \centering
    \includegraphics[width=\linewidth]{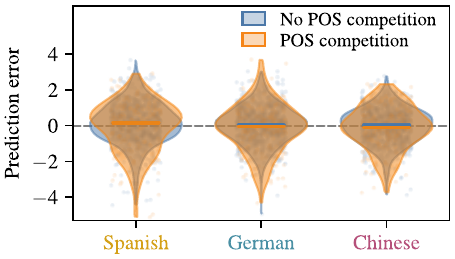}
    \caption{Prediction-error distribution (gold $-$ predicted) by POS competition status across L1 groups (Spanish, German, Chinese). The violin plots compare items without POS competition ($n_{\text{no}}=585$) versus items with POS competition ($n_{\text{yes}}=163$) per language, with horizontal lines marking the median.}
    \label{fig:pos_competition}
\end{figure}

We test whether POS competition affects the prediction error by comparing the prediction errors (gold $-$ predicted) for items with and without POS competition, see \cref{fig:pos_competition}. We find, across all L1s, no significant difference in mean or central tendency (e.g., Spanish: Welch's $t$-test: $p=0.09$, Wilcoxon--Mann--Whitney test: $p=0.36$), indicating that POS competition does not introduce a systematic model-prediction bias. Furthermore, items with POS competition do not differ significantly in gold-label difficulty from items without competition (Welch's $t$-tests and Wilcoxon--Mann--Whitney tests, all $p>0.20$).

However, POS competition has a robust effect on the dispersion of the prediction error. Items with POS competition exhibit substantially higher variance across all L1s, with standard deviations increasing by 24-32\%, see \cref{tab:pos_competition_spread}. This difference is statistically significant according to both Brown--Forsythe and Fligner--Killeen tests ($p < 0.01$), which suggests that POS competition is more difficult to predict for the model, but does not introduce a systematic bias. 

\begin{table}[tbhp]
\centering
\small
\begin{tabular}{cccccc}
\toprule
\textbf{L1} & \textbf{std$_{\text{no}}$} & \textbf{std$_{\text{yes}}$} & \textbf{ratio} & \textbf{BF $p$} & \textbf{FK $p$} \\
\midrule
Spanish & 1.16 & 1.45 & 1.24 & $< 0.01$ & $< 0.01$ \\
German  & 1.04 & 1.37 & 1.32 & $< 0.01$ & $< 0.01$ \\
Chinese & 0.97 & 1.24 & 1.28 & $< 0.01$ & $< 0.01$ \\
\bottomrule
\end{tabular}
\caption{Standard deviation of the prediction error for items with and without POS competition across L1s, with their ratio, and $p$-values from Brown--Forsythe (BF) and
Fligner--Killeen (FK) tests.}
\label{tab:pos_competition_spread}
\end{table}

\begin{figure}[tbhp]
    \centering
    \includegraphics[width=\linewidth]{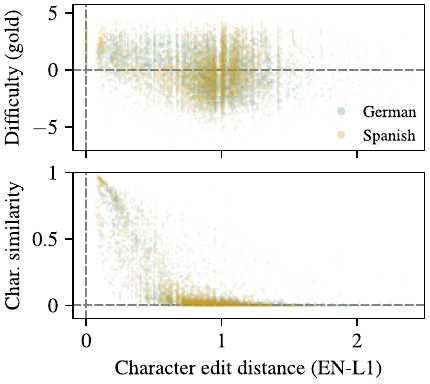}
    \caption{Normalized edit distance between English words and their L1 translations in Spanish and German versus gold-label difficulty (top) and its relationship with character-level cosine similarity (bottom).}
    \label{fig:edit_distance}
\end{figure}

\paragraph{Edit distance.}  
In addition to character-level cosine similarity, we evaluated character edit distance as a measure of orthographic similarity. While edit distance shows a weak-to-moderate negative correlation with gold-label difficulty for Spanish ($r=-0.17$) and German ($r=-0.30$), it is strongly negatively correlated with character-level cosine similarity for Spanish ($r=-0.77$) and German ($r=-0.73$), see \cref{fig:edit_distance}. These two features encode largely the same information, making the inclusion of both redundant. Compared with cosine similarity, edit distance spreads out dissimilar word pairs over a wider range while compressing highly similar forms, making it less informative for distinguishing degrees of cognateness.

\paragraph{L1 frequency.}  

\begin{figure}[tbhp]
    \centering
    \includegraphics[width=\linewidth]{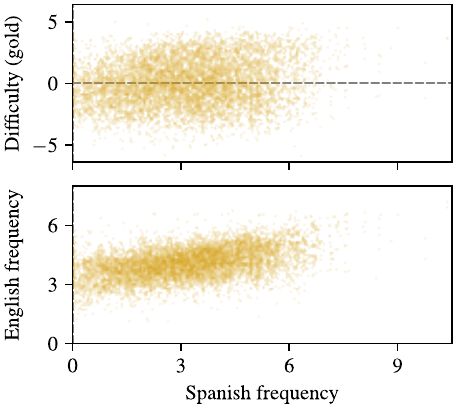}
    \caption{Frequency of L1 (Spanish) source words versus gold-label difficulty (top) and L2 (English) word frequency (bottom).}
    \label{fig:es_frequency}
\end{figure}

Finally, we examine L1 word frequency, which has also been suggested as a predictor of L2 lexical difficulty, under the assumption that more frequent words in a learner's native language may facilitate recognition or acquisition \citep{ellis-2002-frequency, peters-2019-factors}. Using Spanish as an example, \cref{fig:es_frequency} shows that, although higher L1 frequency is associated with lower difficulty ($r = 0.15$), L1 and L2 frequency are themselves strongly positively correlated ($r = 0.44$). As a result, L2 frequency is largely redundant when L1 frequency is already included.

\end{document}